\documentclass[10pt,twocolumn,letterpaper]{article}

\usepackage{cvpr}              %
\usepackage[accsupp]{axessibility}  %

\usepackage{graphicx}
\usepackage{amsmath}
\usepackage{amssymb}
\usepackage{booktabs}
\usepackage{wrapfig}
\usepackage{algorithm}
\usepackage{algorithmic}
\usepackage{multirow}
\usepackage{makecell}

\usepackage{blindtext}
\usepackage{morewrites}
\usepackage{makecell}
\usepackage{pifont}%
\newcommand{\xmark}{\ding{55}}%
\newcommand{\cmark}{\ding{51}}%

\usepackage[pagebackref,breaklinks,colorlinks]{hyperref}

\usepackage[capitalize]{cleveref}
\crefname{section}{Sec.}{Secs.}
\Crefname{section}{Section}{Sections}
\Crefname{table}{Table}{Tables}
\crefname{table}{Tab.}{Tabs.}

\def\cvprPaperID{3} %
\def\confName{CVPR}
\def\confYear{2023}

\begin{document}

\title{Micron-BERT: BERT-based Facial Micro-Expression Recognition \vspace{-5mm}}

\author{Xuan-Bac Nguyen$^{1}$, Chi Nhan Duong$^{2}$, Xin Li$^{3}$, Susan Gauch$^{1}$, Han-Seok Seo$^{4}$, Khoa Luu$^{1}$\\
    $^{1}$ CVIU Lab, University of Arkansas, USA \quad 
	$^{2}$ Concordia University, Canada \\
        $^{3}$ West Virginia University, USA \quad $^{4}$ Dep. of Food Science, University of Arkansas, USA \\
	\tt\small $^{1}$\{xnguyen, sgauch, hanseok, khoaluu\}@uark.edu, $^{2}$\{dcnhan@ieee.org\},  \tt\small$^{3}$xin.li@mail.wvu.edu 
}
\maketitle

\begin{abstract}
Micro-expression recognition is one of the most challenging topics in affective computing. It aims to recognize tiny facial movements difficult for humans to perceive in a brief period, i.e., 0.25 to 0.5 seconds. Recent advances in pre-training deep Bidirectional Transformers (BERT) have significantly improved self-supervised learning tasks in computer vision. However, the standard BERT in vision problems is designed to learn only from full images or videos, and the architecture cannot accurately detect details of facial micro-expressions. This paper presents Micron-BERT ($\mu$-BERT), a novel approach to facial micro-expression recognition. The proposed method can automatically capture these movements in an unsupervised manner based on two key ideas. First, we employ Diagonal Micro-Attention (DMA) to detect tiny differences between two frames. Second, we introduce a new Patch of Interest (PoI) module to localize and highlight micro-expression interest regions and simultaneously reduce noisy backgrounds and distractions. By incorporating these components into an end-to-end deep network, the proposed $\mu$-BERT significantly outperforms all previous work in various micro-expression tasks. $\mu$-BERT can be trained on a large-scale unlabeled dataset, i.e., up to 8 million images, and achieves high accuracy on new unseen facial micro-expression datasets. Empirical experiments show $\mu$-BERT consistently outperforms state-of-the-art performance on four micro-expression benchmarks, including SAMM, CASME II, SMIC, and CASME3, by significant margins. Code will be available at \url{https://github.com/uark-cviu/Micron-BERT}
\end{abstract}
\vspace{-6mm}
\section{Introduction}
\label{sec:intro}

\begin{figure}[!t]
    \centering  
    \includegraphics[width=0.96\columnwidth]{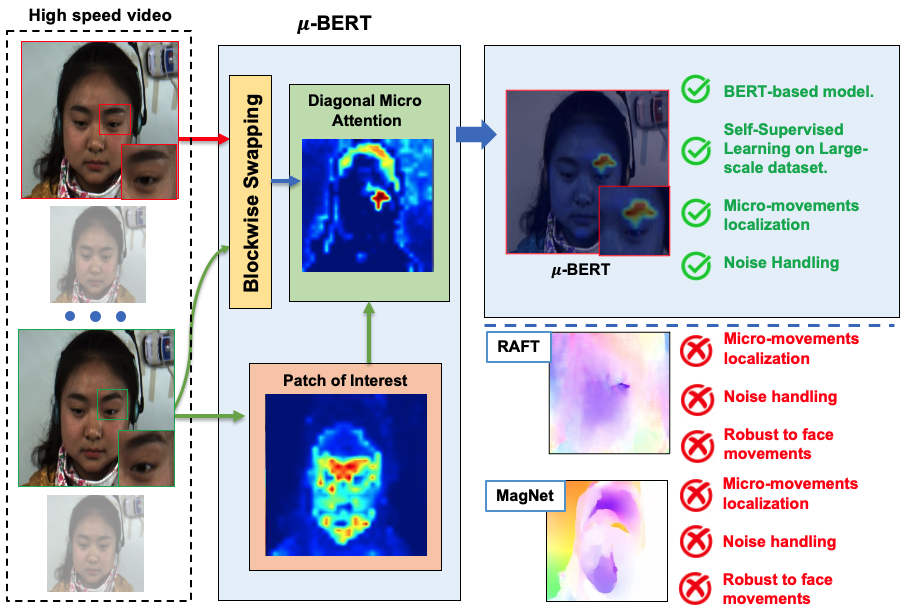}
    \caption{Given two frames from a high-speed video, the proposed $\mu$-BERT method can localize and highlight the regions of micro-movements. \textbf{Best viewed in color.}}
    \label{fig:Fig1}
    \centering
    \vspace{-5mm}
\end{figure}
Facial expressions are a complex mixture of conscious reactions directed toward given stimuli. They involve experiential, behavioral, and physiological elements. Because they are crucial to understanding human reactions, this topic has been widely studied in various application domains \cite{7374704}. In general, facial expression problems can be classified into two main categories, macro-expression, and micro-expression. The main differences between the two are facial expression intensities, and duration \cite{ben2021video}. In particular, macro-expressions happen spontaneously, cover large movement areas in a given face, e.g., mouth, eyes, cheeks, and typically last from 0.5 to 4 seconds. Humans can usually recognize these expressions. By contrast, micro-expressions are involuntary occurrences, have low intensity, and last between 5 milliseconds and half a second. Indeed, micro-expressions are challenging to identify and are mostly detectable only by experts. Micro-expression understanding is essential in numerous applications, primarily lie detection, which is crucial in criminal analysis.

Micro-expression identification requires both semantics and micro-movement analysis. Since they are difficult to observe through human eyes, a high-speed camera, usually with 200 frames per second (FPS) \cite{CASME,samm18,smic13}, is typically used to capture the required video frames. Previous work \cite{9522765} tried to understand this micro information using MagNet \cite{oh18} to amplify small motions between two frames, e.g., onset and apex frames. However, these methods still have limitations in terms of accuracy and robustness.
In summary, the contributions of this work are four-fold:
\begin{itemize}
\vspace{-2mm}
\item A novel Facial Micro-expression Recognition (MER) via Pre-training of Deep Bidirectional Transformers approach (Micron-BERT or $\mu$-BERT) is presented to tackle the problem in a self-supervised learning manner. The proposed method aims to identify and localize micro-movements in faces accurately.
\vspace{-2mm}
\item As detecting the tiny moment changes in faces is an essential input to the MER module, a new \textit{Diagonal Micro Attention} (DMA) mechanism is proposed to precisely identify small movements in faces between two consecutive video frames.
\item A new \textit{Patch of Interest} (POI) module is introduced to efficiently spot facial regions containing the micro-expressions. Far apart from prior methods, it is trained in an unsupervised manner without using any facial labels, such as facial bounding boxes or landmarks.
\vspace{-2mm}
\item The proposed $\mu$-BERT framework is designed in a self-supervised learning manner and trained in an end-to-end deep network. Indeed, it consistently achieves State-of-the-Art (SOTA) results in various standard micro-expression benchmarks, including CASME II \cite{casmeii}, CASME3\cite{casme3}, SAMM\cite{samm18} and SMIC\cite{smic13}. It achieves high recognition accuracy on new unseen subjects of various gender, age, and ethnicity.
\end{itemize}

\section{Related Work}
\label{sec:related_works}

 Generally, prior studies in micro-expression can be divided into two categories, including micro-expression spotting (MES) and micro-expression recognition. 

\noindent \textbf{Micro-Expression Spotting (MES).}
The goal of MES is to determine the specific instant during which a micro-expression occurs. Li et al. \cite{LI2021221} adopted a spatial-channel attention network to detect micro-expression action units.
Tran et al.\cite{tran21} attempted to standardize with the SMIC-E database and an evaluation protocol.
MESNet \cite{wang21} introduced a CNN-based approach with 
a (2+1)D convolutional network, a clip proposal, and a classifier. %

\noindent \textbf{Micro-Expression Recognition (MER).}
The goal of MER tasks is to classify the facial micro-expressions in a video. Ling et al. \cite{9522765} present a new way of learning facial graph representations, allowing these small movements to be seen. Kumar and Bhanu \cite{9523010} 
exploited connections between landmark points and their optical flow patch and achieved %
improvements over state-of-the-art (SOTA) methods for both the CASME II and SAMM. Liu et al. \cite{10.3389/fnbot.2022.922761} presented a new method using transfer learning 
achieved an accuracy of 84.27\% on a composite of three datasets. 
Wang et al. \cite{wang_see_oh_phan_rahulamathavan_ling_tan_li_2016} presented an Eulerian-motion magnification-based approach that highlights these small movements.

\noindent \textbf{Other Work.}
Other research, while not necessarily on MES or MER, is relevant to our approach. %
An advance in video motion magnification is shown in \cite{oh18}, outperforming the SOTA methods in multiple areas. %
This learning-based model can extract filters from data directly rather than rely on ones designed by hand, like the state-of-the-art method. 

\begin{figure*}[t]
    \centering
    \includegraphics[width=0.9\textwidth]{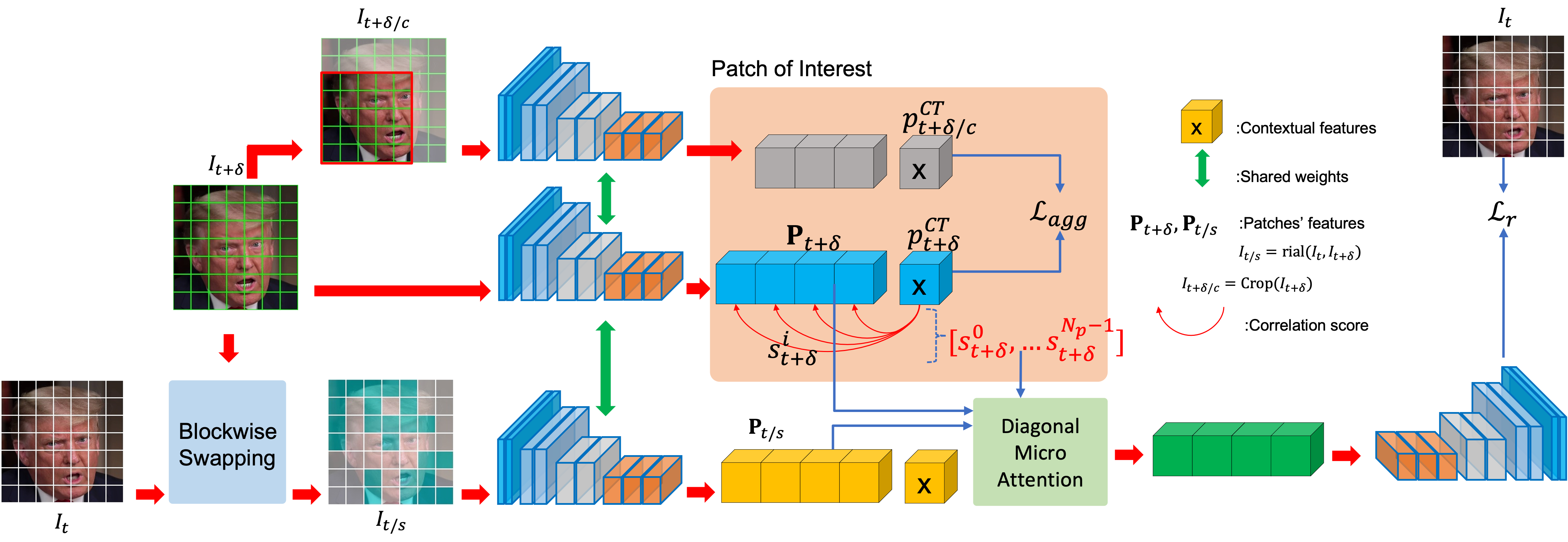}
    \vspace{-2mm}
    \caption{An overview of the proposed $\mu$-BERT approach to facial micro-expression recognition.}
    \vspace{-4mm}
    \label{fig:micro_former}
\end{figure*}

\vspace{-4mm}
\section{BERT Revisited}
\label{sec:basic_bert}

\subsection{BERT in Vision Problems}

Transformers and deep learning have significantly improved results for many tasks in computer vision \cite{dosovitskiy2020image, swin_transformer, swin_transformer_v2, mae, beit, nguyenclusformer, nguyenssl, QUACH2022108646, directformer, rightotalk}. Worth mentioning is Vision Transformer (ViT) \cite{dosovitskiy2020image}, one of the first research efforts at the intersection of Transformers and computer vision. Unlike the traditional CNN network, %
ViT splits an image into a sequence of patches and applies the Transformers-based framework directly. Inspired by the success of BERT in Natural Language Processing (NLP), Bidirectional Encoder representation from Image Transformers (BEiT) \cite{beit} is presented as a self-supervised learning framework in computer vision. In particular, image patches are tokenized using DALL-E \cite{dalle} to the visual tokens. These tokens are then randomly masked before feeding into the transformer backbone. The training objective is to recover the original visual tokens from the corrupted patches. These methods \cite{beit, deit} have marked a remarkable improvement compared to supervised learning methods by leveraging large-scale unlabelled datasets, e.g., ImageNet-1K, ImagNet-21K \cite{ridnik2021imagenet}, to discover semantic information.
\subsection{Limitations of BERT in Vision Problems}
One limitation of using BERT in vision problems is the tokenization step. In the NLP field, a token has precisely one word mapped into it. In vision problems, however, many possible images or patches can share the same token as long as they have the same content. Therefore, designing a BERT model to mask a token and train a prediction model in the missing contexts in computer vision is more challenging than NLP. In addition, the \textit{tokenizer}, i.e., DALL-E \cite{dalle}, is not robust enough to map similar contexts to a token. It yields noise in the tokenization process and affects the overall training performance. He et al., \cite{mae} presented a Masked Auto Encoder (MAE) that utilizes the BERT framework. Instead of tokenizing images, it eliminates patches of an image via a random masking strategy and reconstructs the context of these masked patches to the original content.
Although this method can avoid using the tokenizer, it only considers the context inside an image.
Thus, it does not apply to micro-expression, which requires understanding semantic information from consecutive video frames. In this paper, $\mu$-BERT is presented to address these limitations.

\section{The Proposed $\mu$-BERT Approach}

$\mu$-BERT is designed to model micro-changes of facial texture across temporal dimensions, which is hard to observe by unaided human eyes via a reconstruction process. The proposed $\mu$-BERT architecture, shown in Figure \ref{fig:micro_former}, consists of five main blocks: a \textit{$\mu$-Encoder}, \textit{Patch of Interest (PoI)}, \textit{Blockwise Swapping}, \textit{Diagonal Micro Attention (DMA)}, and a \textit{$\mu$-Decoder}. Given input images $I_t$ and $I_{t+\delta}$, the role of the $\mu$-Encoder is to represent $I_t$ and $I_{t+\delta}$ into latent vectors. Then, Patch of Interest (PoI) constrains $\mu$-BERT to look into facial regions containing micro-expressions rather than unrelated regions such as the background. Blockwise Swapping and Diagonal Micro Attention (DMA) allow the model to focus on facial regions that primarily consist of micro differences between frames. Finally, $\mu$-Decoder reconstructs the output signal back to the determined one. Compared to prior works, $\mu$-BERT can adaptively focus on changes in facial regions while ignoring the ones in the background and effectively recognizes micro-expressions even when face movements occur. Moreover, $\mu$-BERT can also alleviate the dependency on the accuracy of alignment approaches in pre-processing step.

\subsection{Non-overlapping Patches Representation}
In $\mu$-BERT, an input frame $I_t \in \mathbb{R}^{H \times W \times C}$ is divided into a set of several non-overlapping patches $\mathcal{P}_t$ as Eqn. \eqref{eq:PatchRepresentation}.
\begin{equation} \label{eq:PatchRepresentation}
\small
    \mathcal{P}_t = \{p_t^i\}_{i=0}^{N_p-1} \quad\quad |\mathcal{P}_t| = HW/(ps^2) 
\end{equation}
where $H, W, C$ are the height, width, and number of channels, respectively. Each patch $p_t^i$ has a resolution of $ps \times ps$. In our experiments, $H = W = 224$, $C=3$, and $ps=8$. 

\subsection{$\mu$-Encoder}
\label{subsec:encoder}
\begin{figure}[t!]
    \centering
\includegraphics[width=0.3\textwidth,height=0.5\textwidth, keepaspectratio]{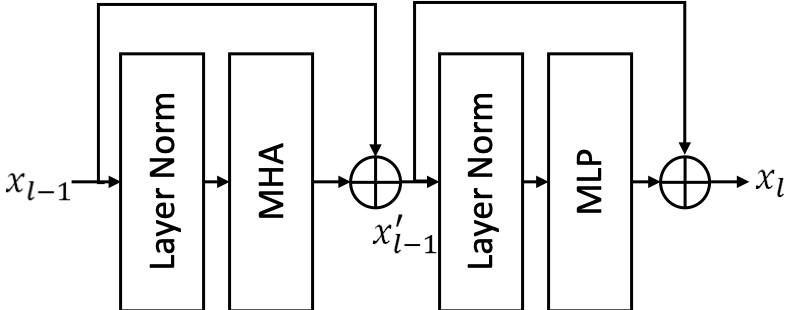}
    \caption{\textbf{Builing block of Encoder and Decoder.} Each block includes Multi-Head Attention (MHA) and Layer Normalization.
    }
    \vspace{-4mm}
    \label{fig:encoder_block}
\end{figure}

Each patch $p_i \in \mathcal{P}_t$ is linearly projected into a latent vector of dimension $d$ denoted as $\mathbf{z}^i_t \in \mathbb{R}^{1\times d}$, with additive fixed positional encoding \cite{vaswani2017attention}. Then, an image $I_t$ can be represented as in Eqn. \eqref{eqn:patch_partition}.
\begin{equation} \label{eqn:patch_partition}
\small
\begin{split}
    \mathbf{Z}_t &= concat \left[\mathbf{z}^0_t, \mathbf{z}^1_t, \dots \mathbf{z}^{N_p - 1}_t\right] \in \mathbb{R}^{N_p \times d} \\ 
    \mathbf{z}^i_t &= \alpha(p^i_t) + \mathbf{e}(i)
\end{split}
\end{equation}
where $\alpha$ and $\mathbf{e}$ are the projection embedding network and positional embedding, respectively. 
Let $\mu$-Encoder, denoted as $\mathcal{E}$, be a stack of continuous blocks. Each block consists of alternating layers of Multi Head Attention (MHA) and Multi-Layer Perceptron (MLP), as illustrated in Figure \ref{fig:encoder_block}. 
The Layer Norm (LN) is employed to the input signal before feeding to MHA and MLP layers, as in Eqn. \eqref{eqn:MHA}.
\begin{equation}
\small
\begin{split}
    \label{eqn:MHA}
    \mathbf{x'}_l &= \mathbf{x}_{l-1} + \textrm{MHA}(\textrm{LN}(\mathbf{x}_{l-1}))  \\
    \mathbf{x}_l &= \mathbf{x'}_l + \textrm{MLP}(\textrm{LN}(\mathbf{x'}_l)) \\
    \mathbf{x}_0 &= \mathbf{Z}_t, \, 1 \leq l \leq L_e
\end{split}
\end{equation}
where $L_e$ is the number of blocks in $\mathcal{E}$.
Given $\mathbf{Z}_t$, 
The output latent vector $\mathbf{P_t}$ is represented as in Eqn. \eqref{eqn:encoder}.
\begin{equation} \label{eqn:encoder}
\small
    \mathbf{P}_t = \mathcal{E}(\mathbf{Z}_t) \quad\quad \mathbf{P}_t \in \mathbb{R}^{N_p \times d}
\end{equation}
\subsection{$\mu$-Decoder} The proposed auto-encoder is designed symmetrically. It means that the decoder part denoted as $\mathcal{D}$, has a similar architecture to the encoder $\mathcal{E}$. Given a latent vector $\mathbf{P}_t$, the decoded signal $\mathbf{Q}_t$ is represented as in Eqn. \eqref{eqn:decoder}.
\begin{equation} \label{eqn:decoder}
    \small
    \mathbf{Q}_t = \mathcal{D}(\mathbf{P}_t) \quad\quad \mathbf{Q}_t \in \mathbb{R}^{N_p \times d}
\end{equation}
We add one more Linear layer to interpolate $\mathbf{Q}_t$ to an intermediate signal $\mathbf{y}_t$ before reshaping it into the image size.
\begin{align}
    \label{eqn:interpolate}
    \mathbf{Q}_t \in \mathbb{R}^{N_p \times d}  & \xrightarrow{\textrm{linear}} \mathbf{y}_t \in \mathbb{R}^{N_p \times ps \times ps \times C} \\  \mathbf{y}_t \in \mathbb{R}^{N_p \times ps \times ps \times C} & \xrightarrow{\textrm{reshape}} \mathbf{y}'_t \in \mathbb{R}^{H \times W \times C} \nonumber
\end{align}
\subsection{Blockwise Swapping}
\label{subsec:copy_paste}

\begin{figure}[t]
    \centering
    \includegraphics[width=0.40\textwidth]{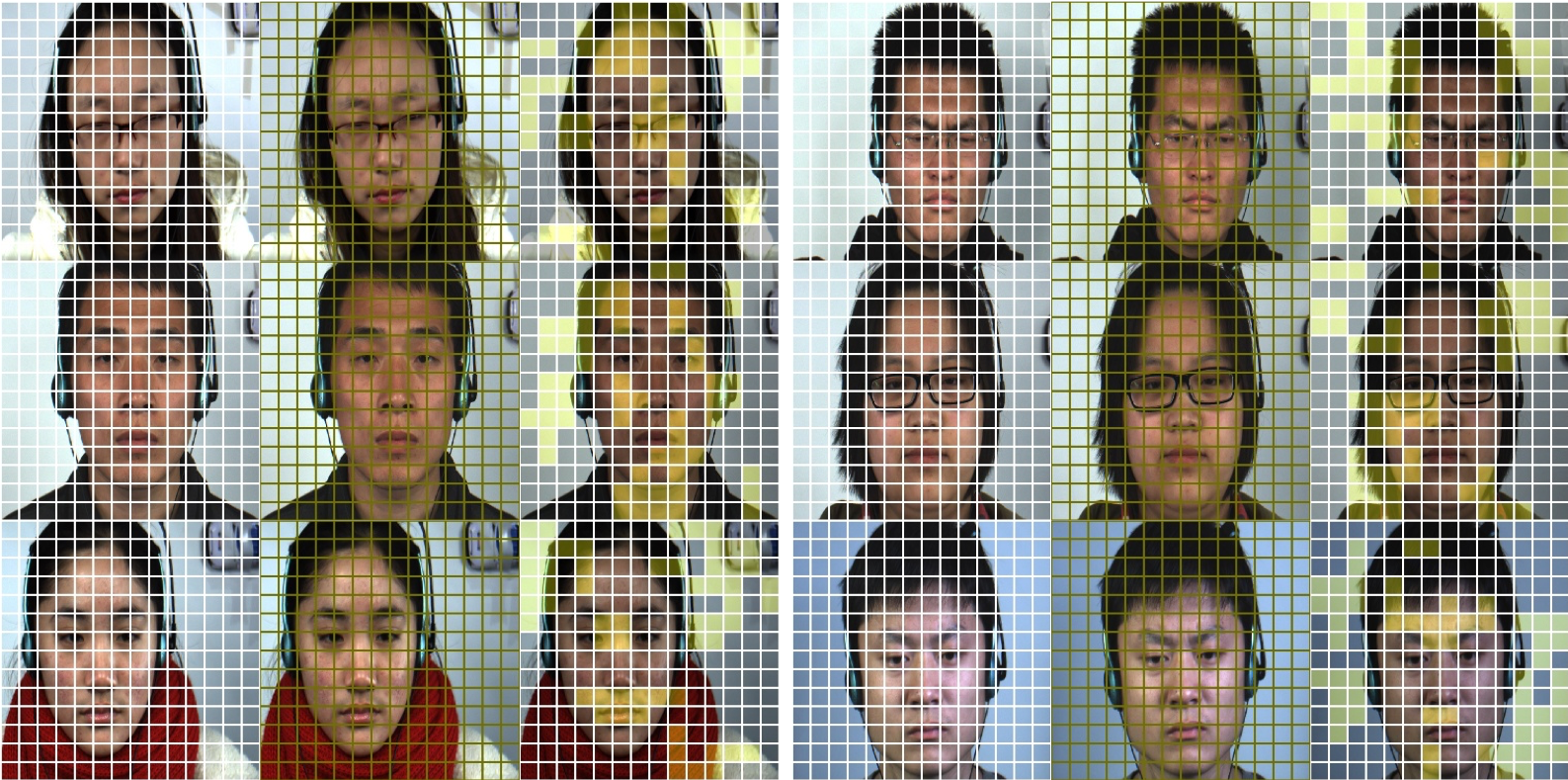}
    \caption{\textbf{Blockwise Swapping.} For each triplet, we present the $I_{t}$ (left), the $I_{t+\delta}$ (middle) and the $I_{t/s}$ (right). The yellow blocks in $I_{t/s}$ represent swapped patches from $I_{t+\delta}$ that are randomly swapped into $I_{t}$. Best viewed in color.}
    \vspace{-4mm}
    \label{fig:copy_paste}
\end{figure}

Given two frames $I_t$ and $I_{t+\delta}$, we realize the fact that:
\begin{equation}
    \label{eq:patch_correlation}
\small
    \lim_{\delta \to 0} s(p^i_t, p^i_{t+\delta}) = 1
\end{equation}
where $p^i_t$ is the $i^{th}$-patch at frame $t$. $s$ denotes a function to measure the similarity between $p^i_t$ and $p^i_{t+\delta}$ where a higher score indicates higher similarity and $0 \leq s(p^i_t, p^i_{t+\delta}) \leq 1$. Given a patch correlation as in Eqn. \eqref{eq:patch_correlation}, 
we propose a \textit{Blockwise Swapping mechanism} to (1) firstly \textit{randomly swap two corresponding blocks} $p^i_t$ and $p^i_{t+\delta}$ between two frames to create a swapped image $I_{t/s}$, and then (2) \textit{enforce the model to spot these changes} and reconstruct $I_t$ from $I_{t/s}$. By doing so, the model is further strengthened in recognizing and restoring the swapped patches. As a result, the learned model can be enhanced by the capability to notice small differences between frames. Moreover, as shown in Eqn. \eqref{eq:patch_correlation}, shorter time $\delta$ causing larger similarity between $I_t$ from $I_{t/s}$ can further help to enhance the robustness on spotting these differences.
The detail of this strategy is described in Algorithm \ref{algo:blockwise_swapping} and Figure \ref{fig:copy_paste}.

\subsection{Diagonal Micro Attention (DMA)}
\label{subsec:diagonal_micro_attention}

\begin{figure}[t]
    \centering
    \includegraphics[width=0.45\textwidth]{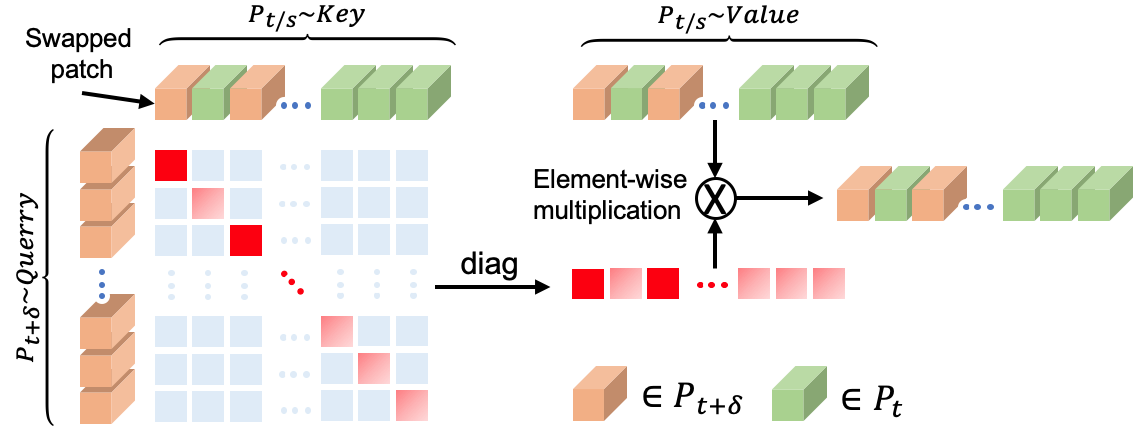}
    \caption{\textbf{Diagonal Micro-Attention (DMA) module.} Diagonal values from the attention map between $\mathbf{P}_{t/s}$ and $\mathbf{P}_{t+\delta}$ are used to rank the importance of each patch in the swapped image.}
    \vspace{-4mm}
    \label{fig:mha_vs_dma}
\end{figure}

As a result of Blockwise Swapping, the image patches $\mathcal{P}_{t/s}$ from $I_{t/s}$ consists of two types, i.e. $p^j_{t/s}$ from $\mathcal{P}_t$ of $I_t$ and $p^i_{t/s}$ from $\mathcal{P}_{t+\delta}$ of $I_{t+\delta}$. Then, the next stage is to learn how to reconstruct $\mathcal{P}_{t}$ from $\mathcal{P}_{t/s}$. 
Since $p^i_{t/s}$ includes all changes between $I_{t}$ and $I_{t/s}$, more emphasis is placed on $p^i_{t/s}$ during reconstruction process. 
Theoretically, the \textit{ground truth} of the index of $p^i_{t/s}$ in $\mathcal{P}_{t/s}$ can be utilized to enforce the model focusing on these swapped patches. However, adopting this information may reduce the learning capability to spot these microchanges. Therefore, a novel attention mechanism named Diagonal Micro-Attention (DMA) is presented to enforce the network automatically focusing on swapped patches $p^i_{t/s}$ and equip it with the ability to precisely spot and identify all changes between images. Notice that these changes may include patches in the background. 
The following section introduces a solution to constrain the learned network focusing on only meaningful facial regions.
The details of DMA are presented in Figure \ref{fig:mha_vs_dma}.
\begin{algorithm}[b]
\centering
\small
\caption{Blockwise Swapping}
\label{algo:blockwise_swapping}
\begin{algorithmic}
\STATE {\bfseries Input:} 
$\quad$ $\mathcal{P}_t$, $\mathcal{P}_{t+\delta}$ image patches ($N_p = h \times w$);  $r_s$: swapping ratio (default: 0.5);  $min\_bs$: minimum block size (default: 16); 
 $min\_ar$: minimum aspect ratio (default: 0.3)
\STATE {\bfseries Output:} Swapped image patches $\mathcal{P}_{t/s}$
\STATE {$\mathcal{P}_{t/s} \gets \mathcal{P}_t$; $N_p \gets |\mathcal{P}_t|$}
\STATE {$c \gets 0$}
\WHILE{$c \leq r_s \times N_p$}

\STATE{$bs \gets$ \textsf{rnd(}$min\_bs$, $r_s \times N_p - c $\textsf{)}} 
\STATE{$ar \gets$ \textsf{rnd($min\_ar$, $1 / min\_ar$\textsf{)}}} %
\STATE{$m, n \gets \sqrt{bs \cdot ar}, \sqrt{bs/ar}$}
\STATE{$p, q \gets$ \textsf{rnd($0$, $h - m$\textsf{)}},  \textsf{rnd($0$, $w - n$\textsf{)}}}
\STATE{$ \forall i \in [p,p+m) , j \in [q,q+n): \; $ \\
\quad \quad $k \gets i \times w + j$ \\ \quad \quad $\mathcal{P}_{t/s}(k) \gets \mathcal{P}_{t+\delta}(k)$}
\STATE{$c \gets c + m \times n$}
\ENDWHILE
\STATE{\textbf{return} $\mathcal{P}_{t/s}$}
\end{algorithmic}
\end{algorithm}
Formally, we construct an attention map $\hat{A}$ between $\mathcal{P}_{t+\delta}$ and $\mathcal{P}_{t/s}$ where the $diag(\hat{A})$ illustrates correlations between two corresponding patches $p^i_{t+\delta}$ and $p^i_{t/s}$. 
From the observation that $ \hat{A}(i, i) > \hat{A}(j, j)$ 
for all $p^i_{t/s} \in \mathcal{P}_{t+\delta}$ and $p^j_{t/s} \in \mathcal{P}_{t}$, $diag(\hat{A})$ can be effectively adopted as  weights indicating important features. 
Full operations of DMA are presented in Eqn. \eqref{eqn:DMA} and Eqn. \eqref{eqn:DMA_Feature}.
\begin{equation}\label{eqn:DMA}
\small
\begin{split}
      \hat{A} &= \textrm{softmax}\left(Q(\mathbf{P}_{t+\delta}) \otimes K(\mathbf{P}_{t/s})^T\right), \sum_{j=0}^{N_p} \hat{A}(i,j) = 1  
\end{split}
\end{equation}
\begin{equation} \label{eqn:DMA_Feature}
\small
\begin{split}
    \mathbf{P}_{dma} &= \textrm{diag}(\hat{A}) \times V(\mathbf{P}_{t/s})  
\end{split}
\end{equation}
where $\times$ denotes the Element-wise multiplication operator.

\subsection{Patch of Interest (POI)}

\begin{figure}[t!]
    \centering
    \includegraphics[width=0.25\textwidth]{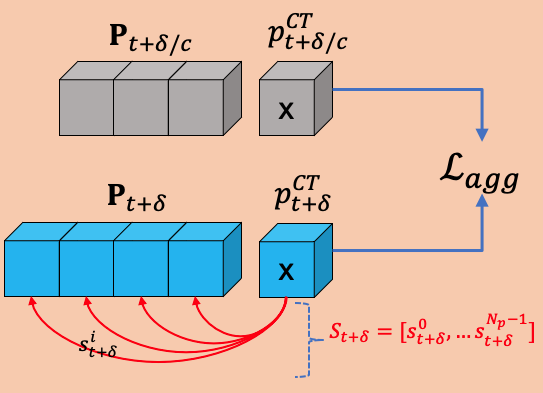}
    \caption{
    \textbf{Patch of Interest (POI) module.} The $\mathbf{P}_{t}$ and $\mathbf{P}_{t+\delta/c}$ are sequence of patch features of $I_{t+\delta}$ and its random cropped version. $p_{t+\delta}^{CT}$ and  $p_{t+\delta/c}^{CT}$ are their corresponding contextual features. }
    \vspace{-5mm}
    \label{fig:poi}
\end{figure}

In Section \ref{subsec:diagonal_micro_attention}, Diagonal Micro-Attention has been introduced to weigh the importance of swapped patches automatically. These swapped patches are randomly produced via Blockwise Swapping, as in Algorithm \ref{algo:blockwise_swapping}. 
In theory, the ideal case is when all swapped patches are located within the facial region only so that the deep network can learn the micro-movements from the facial parts solely and not be distracted by the background. 
In practice, however, we can only identify which parts are selected in the Blockwise Swapping algorithm if the facial regions are available. Thus, the Patch of Interest (POI) is introduced to automatically explore the salient regions and ignore the background patches in an image. Apart from prior methods, the proposed POI leverages the characteristic of self-attention and can be achieved through self-learning without facial labels, such as facial bounding boxes or segmentation masks. The idea of the POI module is illustrated in Figure \ref{fig:poi}.
Thanks to POI, a capability of \textit{automatically focusing on facial regions} is further equipped to the learned model, making it \textit{more robust against facial movements}.

The POI relies on the contextual agreement between the frame $I_{t+\delta}$ and $\textrm{Crop(}I_{t+\delta}\textrm{)}$. Motivated by the BERT framework, we add a Contextual Token $z^{CT}$ to the beginning of the sequence of patches, as in Eqn. \eqref{eqn:patch_partition}, to learn the contextual information in the image. The deeper this token passes through the Transformer blocks, the more information is accumulated from $z^i_t \in \mathcal{P}_t$. 
As a result, $z^{CT}$ becomes a placeholder to store the information extracted from other patches in the sequence and present the contextual information of the image. Let $p^{CT}_{t+\delta}$ and $p^{CT}_{t+\delta/c}$ be the contextual features of frame $I_{t+\delta}$ and its cropped version $\textrm{Crop(}I_{t+\delta}\textrm{)}$ respectively. The agreement loss is then 
defined as in Eqn. \eqref{eqn:agg_loss1}.
\begin{equation} \label{eqn:agg_loss1}
\small
    \mathcal{L}_{agg} = H(p^{CT}_{t+\delta}, p^{CT}_{t+\delta/c})
\end{equation}
where $H$ is the function that enforces $p^{CT}_{t+\delta}$ to be similar to $p^{CT}_{t+\delta/crop}$ so that the model can discover the salient patches. The POI can be extracted from the attention map $A$ at the last attention layer of encoder $\mathcal{E}$. 
In particular, we measure:
\begin{equation} 
\label{eq:POI_score}
\small
\begin{split}
    \mathbf{S}_{t+\delta} = A\left[0,:\right] &= \left[s^0_{t+\delta}, s^1_{t+\delta}, \dots s^{N_p-1}_{t+\delta}\right] 
\end{split}
\end{equation}
where $\sum_{i=0}^{N_p - 1} s^i_{t+\delta} = 1$.
The higher the score $s^i_{t+\delta}$, the richer the patch contains contextual information. Now,  Eqn. \eqref{eqn:DMA_Feature} can be reformulated as in Eqn. \eqref{eqn:DMA_Feature_with_POI}.
\begin{equation} \label{eqn:DMA_Feature_with_POI}
\small
\begin{split}
    \textrm{W} &= \textrm{diag}(\hat{A}) \times {S}_{t+\delta} \\
    \mathbf{P}_{dma} &= \textrm{W} \times V(\mathbf{P}_{t/s})  
\end{split}
\end{equation}

\begin{figure*}[ht]
    \centering
    \includegraphics[width=0.76\textwidth]{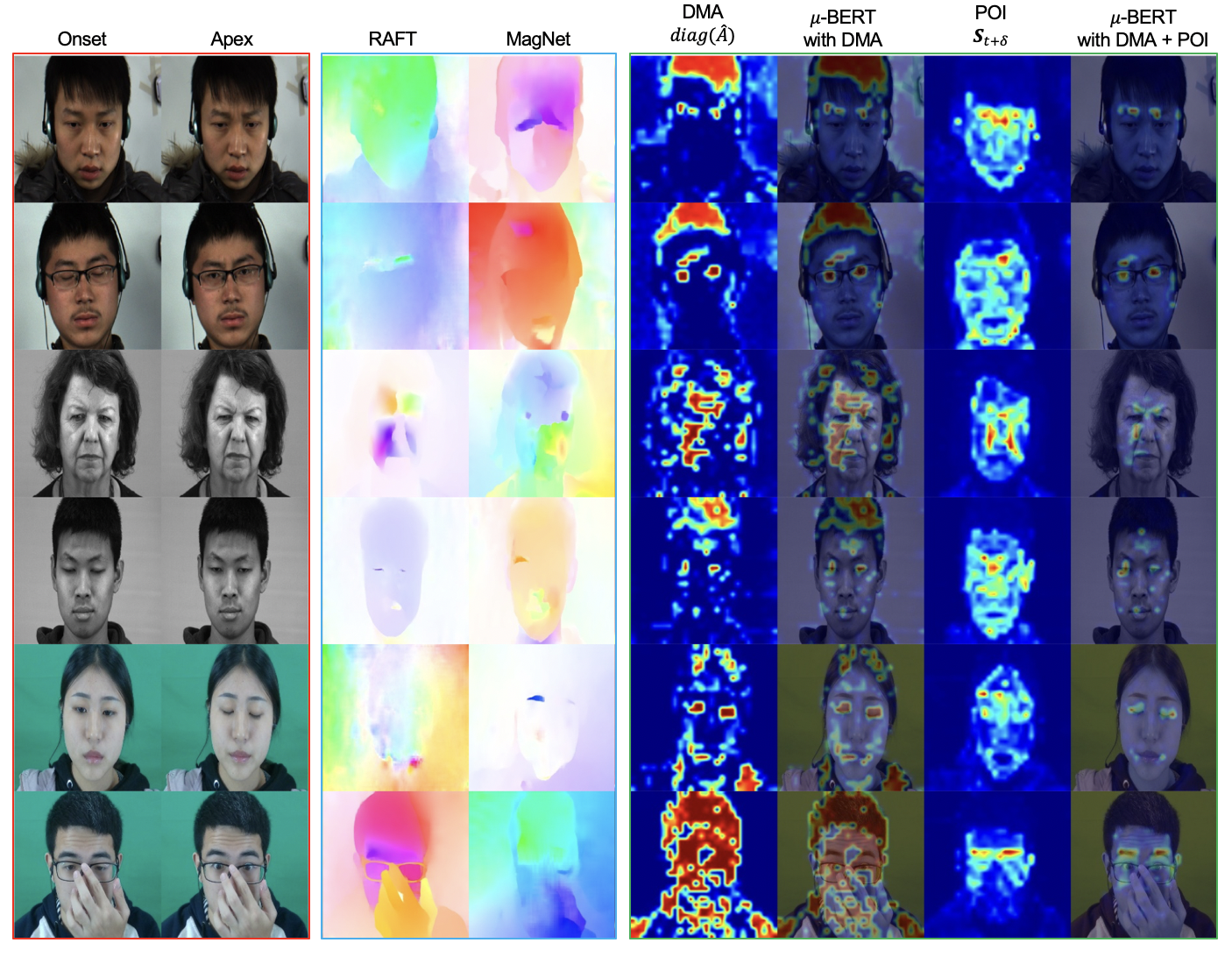}
    \vspace{-2mm}
    \caption{We demonstrate how $\mu$-BERT perceives the tiny differences between two frames. The first two rows are \textit{onset} and \textit{apex} input frames. The third and fourth rows are the results of RAFT and MagNet, respectively. The rest of the rows are our $\mu$-BERT results.}
    \vspace{-5mm}
    \label{fig:micro_comparision}
\end{figure*}

\subsection{Loss Functions}

The proposed $\mu$-BERT deep network is optimized using the proposed loss function as in Eqn. \eqref{eqn:total_loss}.
\begin{equation} \label{eqn:total_loss}
\small
    \mathcal{L} =\gamma \times \mathcal{L}_{r} + \beta \times \mathcal{L}_{agg}
\end{equation}
where $\gamma$ and $\beta$ are the weights for each loss. \\
\textbf{Reconstruction Loss}. 
The output of the decoder $\mathbf{y'}_t$ is reconstructed to the original image $I_t$ using the Mean Square Error (MSE) function.
\begin{equation}
\small
    \mathcal{L}_{r} = \mathrm{MSE}(\mathbf{y'}_t, I_t)
\end{equation}
\noindent \textbf{Contextual Agreement Loss}. MSE is also used to enforce the similarity of contextual features of $I_{t+\delta/crop}$ and $I_{t+\delta}$
\begin{equation}
\small
    \mathcal{L}_{agg} = \mathrm{MSE}(p^{CT}_{t+\delta}, p^{CT}_{t+\delta/crop})
\end{equation}

\section{Experimental Results}

 \subsection{Datasets and Protocols} 
\label{subsec:datasets_and_protocols}
\noindent\textbf{CASME II}\cite{casmeii}. With a 200 fps sampling rate and a facial resolution of $280 \times 340$, CASME II provides 247 micro-expression samples from 26 subjects of the same ethnicity. Labels include apex frames, action units, and emotions.

\noindent\textbf{SAMM}\cite{samm18}. Also, using a 200 fps frame rate and a facial resolution of $400\times400$, SAMM consists of 159 samples from 32 participants and 13 ethnicities. The samples all have emotions, apex frames, and action unit labels.

\noindent\textbf{SMIC}\cite{smic13}. 
SIMC is made up of 164 samples. Lacking apex frame and action unit labels, the samples span 16 participants of 3 ethnicities. The recordings are taken with a resolution of $640\times480$ at 100 fps. 

\noindent\textbf{CASME3}\cite{casme3}. Officially known as CAS(ME)$^3$ provides 1,109 labeled micro-expressions and 3,490 labeled macro-expressions. This dataset has roughly 80 hours of footage with a resolution of $1280\times720$.

\subsection{Micro-Expression Self-Training}
We use all raw frames from CASME3 for self-training except frames of test set. It is important to note that we do not use labels or meta information such as onset, offset, and apex index frames nor labeled emotions. In total, we construct an unlabelled dataset of 8M frames. The images are resized to $224 \times 224$. Then, each image is divided into patches of $8 \times 8$, yielding $N_p = 784$ patches. The temporal index $\delta$ is selected randomly between a lower bound of $5$ and an upper bound of $11$, experimentally. The swapping ratio $r_s$ is selected as $50\%$ of the number of patches being swapped from $I_{t+\delta}$ to $I_{t}$. Each patch is projected to a latent space of $d = 512$ dimensions before being fed into the encoder and decoder. For the encoder and decoder, we keep the same $d$ for all vectors and similar configurations, i.e., $L_e = L_d = 4$. $\mu$-BERT is implemented easily in Pytorch framework and trained by 32 $\times$ A100 GPUs (40G each). The learning rate is set to $0.0001$ initially and then reduced to zero gradually under ConsineLinear \cite{cosinelr} policy. The batch size is set to $64$/GPU. The model is optimized by AdamW \cite{adamw} for $100$ epochs. The training is completed within three days.

\subsection{Micro-Expression Recognition }
\vspace{-2mm}
We leverage the pretrained $\mu$-BERT as an initial weight 
and take the encoder $\mathcal{E}$ and DMA module of $\mu$-BERT as the MER backbone. The input of MER is the \textit{onset} and \textit{apex} frames which correspond to $I_{t}$ and $I_{t+\delta}$ respectively. In Eqn. \eqref{eqn:DMA}, $\mathbf{P}_{dma}$ are the features representing the micro changes and movements between onset and apex frames. They can be effectively adopted for recognizing micro-expressions. 
We adopt the standard metrics and protocols of MER2019 challenge \cite{megc_2019} with the unweighted F1 score $UF1 = \frac{1}{C} \sum_{i=0}^{C-1}\frac{2 \times TP_i}{TP_i + FP_i + FN_i}$ 
and accuracy $UAR = \frac{1}{C} \sum_{i=0}^{C-1}\frac{TP_i}{N_i}$, where $C$ is the number of MEs, $N_i$ is the total number of $i^{th}$ ME in the dataset. %
Leave-one-out cross-validation (LOOCV) scheme is used for evaluation.

\subsection{Results}
Our proposed $\mu$-BERT shows a significant improvement over prior methods and baselines, as shown in Table \ref{tab:result_CASME3} on the CASME3. %
Tested using 3, 4, and 7 emotion classes, $\mu$-BERT achieves double-digit gains over the compared methods in each category. In the case of 3 emotion classes, $\mu$-BERT achieved a 56.04\% UF1 score and 61.25\% UAR, compared to RCN-A's \cite{rcn_a} 39.28\% UF1 and 38.93\% UAR. For 4 emotion classes, $\mu$-BERT outperforms Baseline (+Depth) \cite{casme3} 47.18\% to 30.01\% for UF1 and 49.13\% to 29.82\% for UAR. Large gains over Baseline (+Depth) \cite{casme3} are seen in the case of 7 emotion classes, where $\mu$-BERT attains UF1 and UAR scores of 32.64\% and 32.54\% respectively, compared to 17.73\% and 18.29\% for the baseline. 

Table \ref{tab:result_CASMEII} details results for CASMEII. $\mu$-BERT shows improvements over all other methods. For three categories, it achieves a UF1 of 90.34\%  and UAR of 89.14\%, representing 3.37\% and 0.86\% increases over the prior leading method (OFF-ApexNet \cite{gan2019off}), respectively. Similar improvement is seen in five categories: a 4.83\% over TSCNN \cite{song2019recognizing} in terms of UF1 and a 0.89\% increase over SMA-STN \cite{liu2020sma} for UAR.
Similarly, $\mu$-BERT performs competitively with other methods on the SAMM as seen in Table \ref{tab:result_SAMM}. 
Using 5 emotion classes, $\mu$-BERT outperforms MinMaNet\cite{xiamicro} by a large margin in terms of UF1 (83.86\% vs 76.40\%) and UAR (84.75\% vs 76.70\%), respectively. The performance of $\mu$-BERT on SMIC is compared against several others in Table \ref{tab:result_SMIC}. $\mu$-BERT outperforms others with a 7.5\% increase in UF1 to 85.5\% and a 3.97\% boost in UAR to 83.84\%.

\begin{table}[t]
\small
\renewcommand{\arraystretch}{1.3}
\centering 
\caption{MER on the CASME3 dataset.}
\vspace{-2mm}
\resizebox{1.\columnwidth}{!}{
\begin{tabular}{l c c c}
\bottomrule[1.5pt]
\small
Method & $\#$ Classes & UF1 (\%) & UAR(\%) \\ \bottomrule[1.5pt]
FR~\cite{FR} & 3 & 34.93 & 34.13 \\ %
STSTNet~\cite{ststnet} & 3 & 37.95 & 37.92 \\ %
RCN-A~\cite{rcn_a} & 3 & 39.28 & 38.93 \\ %
\textbf{$\mu$-BERT (ours)} & 3 & \textbf{56.04} & \textbf{61.25} \\ %
\hline
Baseline~\cite{casme3} & 4 & 29.15 & 29.10 \\ %
Baseline (+Depth)~\cite{casme3} & 4 & 30.01 & 29.82 \\ %
\textbf{$\mu$-BERT (ours)} & 4 & \textbf{47.18} & \textbf{49.13} \\ %
\hline
Baseline~\cite{casme3} & 7 & 17.59 & 18.01 \\ %
Baseline(+Depth)~\cite{casme3} & 7 & 17.73 & 18.29 \\ %
\textbf{$\mu$-BERT (ours)} & 7 & \textbf{32.64} & \textbf{32.54} \\
\hline
\end{tabular}
}
\vspace{-2mm}
\label{tab:result_CASME3}
\end{table}

\begin{table}[t]
\renewcommand{\arraystretch}{1.3}
\centering 
\small
\caption{ MER on CASME II dataset. }
\vspace{-2mm}
\resizebox{1.\columnwidth}{!}{
\begin{tabular}{l c c c}
 \bottomrule[1.5pt]
 \small
 Method & $\#$ Classes & UF1 (\%) & UAR (\%)\\
 \bottomrule[1.5pt]

LR-GACNN \cite{Kumar_2021_CVPR} 
& 5 & 70.90 & 81.30\\ %

AMAN\cite{wei2022novel} 
& 5 & 71.00 & 75.40 \\ %

Graph-TCN~\cite{lei2020novel} 
& 5 & 72.46 & 73.98\\ %

DSTAN\cite{wang2021micro} 
& 5 & 73.00 & 75.00 \\ %

GEME  \cite{nie2021geme} 
& 5 & 73.54 & 75.20\\ %

MiMaNet \cite{xiamicro}
& 5 & 75.90 & 79.90\\ %

SMA-STN~\cite{liu2020sma} 
& 5 & 79.46 & 82.59\\ %

TSCNN~\cite{song2019recognizing} 
& 5 & 80.70 & 80.97\\ %

\textbf{$\mu$-BERT (ours)} 
& 5 & \textbf{85.53} & \textbf{83.48} \\ \hline

STSTNet~\cite{liong2019shallow} 
& 3 & 83.82 & 86.86\\ %

OFF-ApexNet \cite{gan2019off} 
& 3 & 86.97 & 88.28\\ %

MAE \cite{mae}
& 3 & {88.03} & {87.28} \\

\textbf{$\mu$-BERT (ours)} 
& 3 & \textbf{90.34} & \textbf{89.14}

\\ \hline %

\end{tabular}

}
\vspace{-4mm}
 \label{tab:result_CASMEII}
\end{table}

\begin{table}[t]
\renewcommand{\arraystretch}{1.3}
\centering 
\small
\caption{ MER on SAMM dataset. }
\vspace{-2mm}
\resizebox{1.\columnwidth}{!}{
\begin{tabular}{l c c c}
 \bottomrule[1.5pt]
 \small
Method 
& $\#$ Classes & UF1 (\%) & UAR (\%)\\
\bottomrule[1.5pt]

AMAN\cite{wei2022novel} 
& 5 & 67.00 & 68.85\\ %

SMA-STN~\cite{liu2020sma}
& 5 & 70.33 & 77.20\\ %

GRAPH-AU \cite{Lei_2021_CVPR} 
& 5 & 70.45 & 74.26\\ %

MTMNet \cite{xia2020learning} 
& 5 & 73.60 & 74.10\\ %

MiMaNet \cite{xiamicro} 
& 5 & 76.40 & 76.70\\ %

MAE \cite{mae}
& 5 & {80.40} & {88.98} \\

\hline
\textbf{$\mu$-BERT (ours)} 
& 5 & \textbf{83.86} & \textbf{84.75} \\

\hline

\end{tabular}
}
\vspace{-2mm}
 \label{tab:result_SAMM}
\end{table}

\begin{table}[t]
\renewcommand{\arraystretch}{1.3}
\centering 
\small
\caption{MER on SMIC dataset.}
\vspace{-2mm}
\resizebox{1.\columnwidth}{!}{
\begin{tabular}{l c c c}
 \bottomrule[1.5pt]
 \small
Method 
& $\#$ Classes & UF1 (\%) & UAR (\%)\\
 \bottomrule[1.5pt]

DIKD~\cite{sun2020dynamic} 
& 3 & 71.00 & 76.06
\\ %

TSCNN~\cite{song2019recognizing} 
& 3 & 72.36 & 72.74\\ %

MTMNet \cite{xia2020learning}  
& 3 & 74.40 & 76.00\\ %

AMAN\cite{wei2022novel}
& 3 & 77.00 & 79.87\\ %

MiMaNet \cite{xiamicro} 
& 3 & 77.80 & 78.60\\ %

DSTAN\cite{wang2021micro}
& 3 & 78.00 & 77.00 \\ %

MAE \cite{mae}
& 3 & {81.86} & {80.82} \\

\hline

\textbf{$\mu$-BERT (ours)} 
& 3 & \textbf{85.50} & \textbf{83.84} \\

\hline
\end{tabular}
}

\vspace{-4mm}
 \label{tab:result_SMIC}
\end{table}

On the composite dataset, $\mu$-BERT again outperforms other methods (Table \ref{tab:result_MER2019}). Attaining a UF1 score of 89.03\% and UAR of 88.42\%, $\mu$-BERT realizes 0.73\%, and 0.82\% gains over previous best MiMaNet \cite{xiamicro}, respectively.
Table \ref{tab:ablationstudy} shows the impact of  DMA and POI on CASME3. 
Our method gives more modest gains of approximately 2\% in both metrics.
A greater improvement is seen with DMA, where UF1 and UAR increase by another 2-4\%. Significant improvement from $\mu$-BERT is seen when adopting both modules, 
with a UF1 of 32.64\% and UAR of 32.54\%, representing roughly 10\% gains over previous methods.

\subsection{How $\mu$-BERT perceives micro-movements} 

To understand the micro-movements between two frames, the onset and apex frames are inputs for $\mu$-BERT. These frames represent the moments that the micro-expression starts and is observed. We measure $\textrm{diag}(\hat{A})$ (Subsection \ref{subsec:diagonal_micro_attention}) and $\mathbf{S}_{t+\delta}$ (Eqn \eqref{eq:POI_score}) values to identify which regions contain small movements between two frames. Comparisons of $\mu$-BERT with RAFT \cite{raft}, i.e., optical flow-based method and MagNet \cite{oh18} are also conducted as in Fig. \ref{fig:micro_comparision}. 
The third and fourth columns in Fig \ref{fig:micro_comparision} show the results of RAFT \cite{raft}, and MagNet \cite{oh18} on spotting the micro-movements, respectively. While RAFT is an optical flow-based method, 
MagNet 
amplifies small differences between the two frames. These methods are sensitive to the environment (e.g., lighting, illuminations). Thus, noises in the background still exist in their outputs. In addition, neither RAFT nor MagNet understand semantic information in the frame and distinguish changes inside facial or background regions.
Meanwhile, $\mu$-BERT shows its advantages in perceiving micro-movements via distinguishing the facial regions and spotting the micro-expressions. In particular, the attention map in the fifth column, in Fig. \ref{fig:micro_comparision} illustrates the micro-differences between onset and apex frames. The higher contrast represents the higher chance of small movements in these regions. With the POI module, $\mu$-BERT can automatically figure out the informative patches and ignore the background ones.
Then, with %
DMA module, $\mu$-BERT, can detect and localize which corresponding patches/regions contain tiny movements. 
As shown in the seventh column, attention maps represent the most salient regions in the image. By empowering DMA and POI, $\mu$-BERT effectively identifies micro-movements within facial regions, as demonstrated in the last column.
\begin{table}[t]
\renewcommand{\arraystretch}{1.3}
\centering 
\small
\caption{MER on the Composite dataset (MECG2019).}
\vspace{-2mm}
\resizebox{1.\columnwidth}{!}{
\begin{tabular}{l c c c}
\bottomrule[1.5pt]
\small
Method 
& $\#$ Classes & UF1 (\%) & UAR(\%) \\
\bottomrule[1.5pt]

Dual-Inception~\cite{zhou2019dual} 
& 3 & 73.22 & 72.78\\ %

FR \cite{zhou2021feature} 
& 3 & 78.38 & 78.32\\ %

NMER ~\cite{liu2019neural} 
& 3 & 78.85 & 78.24   \\ %

GRAPH-AU \cite{Lei_2021_CVPR} 
& 3 & 79.14 & 79.33\\ %

ICE-GAN~\cite{yu2020ice} 
& 3 & 84.50 & 84.10\\ %

BDCNN \cite{chen2022block} 
& 3 & 85.09 & 85.00 \\ %

moment \cite{xia2020learning} 
& 3 & 86.40 & 85.70\\ %

MiMaNet \cite{xiamicro} 
& 3 & 88.30 & 87.60\\ %

MAE \cite{mae}
& 3 & {88.50} & {87.40} \\

\hline

\textbf{$\mu$-BERT (ours)} 
& 3 & \textbf{89.03} & \textbf{88.42}

\\ \hline %
\end{tabular}
}
\vspace{-4mm}
\label{tab:result_MER2019}
\end{table}

\vspace{-1mm}
\subsection{Ablation studies}
\vspace{-1mm}

\noindent This section compares $\mu$-BERT against other self-supervised learning (SSL) methods on the MER task. CASME3 is used for experiments since it has many unlabelled images to demonstrate the power of SSL methods. We also analyze the essential contributions of Diagonal Micro-Attention (DMA) and Patch of Interest (POI) modules. Finally, we illustrate the robustness of $\mu$-BERT pretrained on CASME3 on unseen datasets and domains. 

\noindent \textbf{Comparisons with self-supervised learning methods}. We utilize the encoder and decoder parts of $\mu$-BERT (\textbf{without} DMA and POI) to train previous SSL methods (MoCo V3 \cite{moco}, BEIT \cite{beit}, MAE \cite{mae}) and then continue learning the MER task on the large-scale database CASME3 \cite{casme3}. Overall results are shown in Table \ref{tab:ablationstudy}. It is expected that ViT-S achieves the lowest performance for UF1 and UAR as ImageNet and Micro-Expression are two different domains. Three self-supervised methods (MoCo V3, BEIT, and MAE) got better results when they were pretrained on CASME before fine-tuning to the recognition task. Compared to ViT-S, these SSL methods gain remarkable performance. Especially, MAE \cite{mae} achieves 3.5\% and 2\% up on UF1 and UAR compared to ViT-S, respectively.

\noindent
\textbf{The role of Blockwise Swapping.} Our basic setup of $\mu$-BERT (denoted as MB1) is employed to train in an SSL manner. It is noted that only Blockwise Swapping is involved, and it does not contain either DMA or POI. Compared to MAE, MB1 outperforms MAE by 2\% in both UF1 and UAR, approximately. The reasons are: (1) Blockwise Swapping enforces the model to learn local context features inside an image, i.e., $I_t$, and (2) It helps the network to figure out micro-disparities between two frames $I_t$ and $I_{t+\delta}$.

\noindent
\textbf{The role of DMA.} This module is the guide to tell the network where to look and which patches to focus. By doing so, the $\mu$-BERT gets more robust knowledge of micro-movements between two frames. For this reason, the network (denoted as MB2) achieves 2\% on UF1 and a significant 4\% gain on UAR compared to MB1. 

\noindent
\textbf{The role of POI.} Since MB1 are sensitive to background noise, the micro-disparities features $\mathbf{P}_{dma}$ might contain unwanted features coming from the background. The POI is designed as a filter that only lets the typical interesting patches belonging to the subject go through and preserves the micro-movement features only. The improvements of up to 6\% compared to MB2 demonstrate the important role of POI in $\mu$-BERT for micro-expression tasks. Qualitative results demonstrated in Supplementary Material can further emphasize the advantages of POI in assisting the network to be 
robust against facial movements.

\begin{table}[t]
\small
\centering
\caption{MER performance on CASME3 by different self-supervised methods and various settings of $\mu$-BERT}
\vspace{-2mm}
\label{tab:ablationstudy}
\resizebox{1.\columnwidth}{!}{
    \begin{tabular}{@{}ccccll@{}}
    \bottomrule[1.5pt]
    Method & Pre-train & DMA & POI & UF1 & UAR \\
    \bottomrule[1.5pt]
    ViT-S \cite{dosovitskiy2020image} & ImageNet & \xmark & \xmark & 20.34 & 18.76 \\
    MoCo V3 - R50 \cite{moco} & CASME3 & \xmark & \xmark & 19.12 & 17.36 \\
    MoCo V3 - R101 \cite{moco} & CASME3 & \xmark & \xmark & 20.14 & 18.52 \\
    MoCo V3 \cite{moco} & CASME3 & \xmark & \xmark & 22.13 & 19.34 \\
    BEIT \cite{beit} & CASME3 & \xmark & \xmark & 23.54 & 19.89 \\
    MAE \cite{mae} & CASME3 & \xmark & \xmark & 23.86 & 20.87 \\
    \hline
    $\mu$-BERT (MB1) & CASME3 & \xmark & \xmark & 25.27 & 22.96 \\
    $\mu$-BERT (MB2) & CASME3 & \cmark & \xmark & 27.35 & 26.18 \\
    $\mu$-BERT (MB3) & CASME3 & \cmark & \cmark & \textbf{32.64} & \textbf{32.54} \\
    \hline
    \end{tabular}
}
\vspace{-4mm}
\end{table}

\vspace{-3mm}
\section{Conclusions and Discussions}
\vspace{-1mm}

Unlike a few concurrent research on micro-expression, we move forward and study how to explore BERT pretraining for this problem. In our proposed $\mu$-BERT, we presented a novel Diagonal Micro Attention (DMA) to learn the micro-movements of the subject across frames. The Patch of Interest (POI) module is proposed to guide the network, focusing on the most salient parts, i.e., facial regions, and ignoring the noisy sensitivities %
from the background. Empowered by the simple design of $\mu$-BERT, SOTA performance on micro-expression recognition tasks is achieved in four benchmark datasets. Our perspective will inspire more future study efforts in this direction.

\noindent
\textbf{Limitations}. We demonstrated the efficiency of the POI module in removing noise in the background, which is sensitive to lighting and illumination. However, suppose any facial parts, e.g., the forehead, are affected by lighting conditions while there are no movements. In that case, these noisy factors might also be included as micro-difference features. The robustness with different lighting conditions will be left as our future works.

\vspace{-4mm}
\paragraph{Acknowledgement} This work is supported by Arkansas Biosciences Institute (ABI) Grant, NSF WVAR-CRESH and NSF Data Science, Data Analytics that are Robust and Trusted (DART). We also acknowledge the Arkansas High-Performance Computing Center for providing GPUs.

{\small
\bibliographystyle{ieee_fullname}
\bibliography{egbib}
}

\end{document}